# Hierarchical Classification of Transversal Skills in Job Ads Based on Sentence Embeddings

**Florin Leon[1], Marius Gavrilescu[1], Sabina-Adriana Floria[1], Alina-Adriana Minea[2]**

[1] Faculty of Automatic Control and Computer Engineering
[2] Faculty of Materials Science and Engineering
"Gheorghe Asachi" Technical University of Iași, Romania
Emails: florin.leon@academic.tuiasi.ro, marius.gavrilescu@academic.tuiasi.ro,
sabina-adriana.floria@academic.tuiasi.ro, alina-adriana.minea@academic.tuiasi.ro

**Abstract**

This paper proposes a classification framework aimed at identifying correlations between job ad requirements and transversal skill sets, with a focus on predicting the necessary skills for individual job descriptions using a deep learning model. The approach involves data collection, preprocessing, and labeling using ESCO (European Skills, Competences, and Occupations) taxonomy. Hierarchical classification and multi-label strategies are used for skill identification, while augmentation techniques address data imbalance, enhancing model robustness. A comparison between results obtained with English-specific and multi-language sentence embedding models reveals close accuracy. The experimental case studies detail neural network configurations, hyperparameters, and cross-validation results, highlighting the efficacy of the hierarchical approach and the suitability of the multi-language model for the diverse European job market. Thus, a new approach is proposed for the hierarchical classification of transversal skills from job ads.

## 1. Introduction

The field of text classification, a fundamental subdomain within the natural language processing (NLP) field of machine learning (ML), has witnessed a remarkable evolution in recent years. With the exponential increase in textual data generated across various domains, the need for effective text classification methods has become increasingly pressing. Text classification is the task of assigning predefined labels or categories to textual documents based on their content. This task holds immense importance across various industries and applications, including but not limited to sentiment analysis, spam detection, content recommendation, and news classification. The ability to automatically organize and categorize large volumes of text can streamline information retrieval, enhance decision-making processes, and enable efficient data management.

Traditional text classification methods rely on well-established techniques such as term frequency - inverse document frequency (TF-IDF) representations and traditional ML algorithms. TF-IDF measures the importance of each term within a document relative to a corpus of documents, providing a numerical representation of textual data. Classic ML algorithms, such as k-nearest neighbors, decision trees, naïve Bayes or random forest, process the TF-IDF vectors to identify patterns and relationships among terms, and have been successfully applied to text classification tasks.

While classic methods yielded commendable results, the emergence of deep learning (DL) has brought a new era of text classification. DL models, i.e., neural networks, have demonstrated unprecedented capabilities in handling the complexity and nuances of natural language. One of the key breakthroughs in DL for text classification lies in the use of word and sentence embeddings that represent words as vectors in high-dimensional spaces, capturing semantic relationships between them. Sentence embeddings extend this idea to encode entire sentences or documents into vector-based representations.

DL models use these embeddings to learn complex patterns and contextual information within text data. Recurrent neural networks, e.g., long-short term memory (LSTM) models, and more advanced architectures such as those based on Transformers have achieved state-of-the-art performance across a wide range of text classification tasks.

The motivation behind employing text classification for the identification of skills from job advertisements is rooted in the ever-evolving job market dynamics and the imperative need for efficient workforce matching. In today's fast-paced and highly competitive job market, where the demand for specific skills is continually changing, exploiting the power of text classification offers several significant advantages.

First, it allows for precise talent matching. Job seekers possess a diverse range of skills, and employers have specific skill requirements for their open positions. Text classification ensures that the right individuals with the exact skill sets are efficiently matched with job opportunities that necessitate those particular competences. This results in reduced skill mismatches, higher job satisfaction, and increased productivity.

Moreover, the job market is a dynamic entity characterized by rapid skill turnover due to technological advancements and evolving industry needs. Text classification offers the ability to perform real-time analysis of job descriptions and skill requirements. By staying up-to-date with the latest trends and demands, organizations can rapidly adapt, ensuring that their workforce remains competitive and aligned with market needs.

Furthermore, text classification optimizes resource allocation for human resources departments and job search platforms. It automates the labor-intensive process of scanning and categorizing skills from large numbers of job listings or CVs, thus saving valuable time and effort.

Lastly, another compelling motivation is the identification of skill gaps. For job seekers, text classification helps individuals pinpoint areas where they may lack necessary skills for a specific role. This knowledge empowers them to proactively seek out relevant training or education, fostering lifelong learning and career development.

Our contribution involves the development of neural network models designed for the classification of job ads based on the ESCO taxonomy. To accomplish this, we create our own dataset, comprising manually labeled job ads that reference the skills mentioned in the respective text. Additionally, our classifiers consider the hierarchical structure outlined in the ESCO platform, wherein skills are categorized into two levels: the higher level comprises more broadly-phrased skills with a wider context, while the lower level encompasses sets of subskills for each top-level skill, providing a more detailed specification.

The rest of the paper is structured as follows. In Section 2, we include a short survey of the related literature. Section 3 details the methodology, including data preprocessing and the proposed hierarchical skill classification. In Section 4, we show the experimental results, and in Section 5 we present the conclusions of this work.

## 2. Related Work

In general, the biggest obstacles in classifying job ads and identifying relevant skills can be summarized as: the availability of data, particularly pre-labeled sets of job ads and corresponding skills that can be used for training supervised classifiers; the unstructured nature of many job ads, which feature a broad range of formats, phrasings, different manners of expressing similar requirements, and vagueness when formulating job requirements. These aspects cause the problem space to increase disproportionately compared to the available data, making it tedious and difficult to properly preprocess and standardize the training data, as well as to develop reliable classification models. Consequently, skill identification is a multifaceted topic addressed in a variety of ways in the related literature.

One of the more common and reliable methods for skill identification is skill counting. Manual skill counting relies on expert readers to identify relevant skills in job ads. This can be achieved without a knowledge base or using an already available skill set. The drawback of such an approach is that it is time-consuming and tedious, however the availability of a prior list of skills to choose from makes the task easier for expert staff. The process can also be automated if a skill base is available. The competences can be identified either by Boolean indexing, or using a simple feature as a word weight, which indicates the importance of the skill's phrasing. A common weighing method makes use of metrics such as TF/IDF [1].

Early work in the direction of skill identification involved finding exact matches of skill labels from existing skill bases in job ad phrasing. Such methods are simple to implement and rely on searching for keywords or keyword combinations [2-4]. In absence of a skill base, skill counting mainly relies on the assessment of expert annotators. The topic of manual or semi-automated content analysis has been a subject of thorough research, since it has the distinctive advantage of also performing a qualitative search alongside a quantitative one. Multiple studies employ content analysis for identifying job market needs, though the general consensus is that manual annotation, while often more reliable, is time consuming and ineffective for the systematic analysis of large bodies of text [5-8].

Multiple authors handle the skill search task by treating it as a topic modeling problem. In topic modeling, the main themes of a text are learned in an unsupervised manner, by determining and analyzing word distributions. The works that employ topic modeling algorithms identify the most required skills among the topics of the job ad. A common approach in this sense is to find the formulation of relevant skills in the most frequent keywords of the identified topics. In this context, the authors of [9] use latent semantic analysis (LSA) to carry out skill identification from job ads. LSA involves transforming the job ad set into a term matrix. This matrix is then subjected to dimensionality reduction via singular value decomposition, which results in sets of highly-correlated keywords and documents. The phrases formed by these keywords are the topics identified in the texts, which are further subjected to data analysis techniques and expert evaluation. Similar works are by [10-12], where latent Dirichlet allocation (LDA) is used to identify popular keywords in job ads. Each document is transformed into a probability distribution of topics and each topic is treated as a probability distribution of words, all sharing a common Dirichlet prior [13].

Word embeddings have become increasingly popular in generating and classifying text. Similar to word embeddings, skill embedding methods generate vector representations of skill keywords, such that similar skills have high similarity in the corresponding vector space. The aim of most works is to develop embedding spaces that work with simple similarity metrics, such as cosine or Jaccard similarities. In [14-15], the authors employ the Word2vec model [16] to derive vector representations for skill contexts. Subsequently, these vectors serve as inputs for a clustering algorithm, facilitating the grouping of aggregated contexts in clusters.

In [17] the authors use Word2vec embeddings to assess the similarity between skills mentioned in job ads and in professional standards. These standards encompass a set of principles, ethics, and behaviors obligatory for members of a specific profession. The training of Word2vec on a job ad corpus enables the model to learn contextual information from job ads, facilitating the clustering of skills based on their presence in the job ad texts.

The authors of [18] first identify skills explicitly, then infer implicit skills from job ads using Doc2vec [19]. The inference process involves identifying similar job descriptions that share common features such as location or company, assuming that they would also share similar skills. Implicit skills, in this context, are those not explicitly stated but considered important for a given position. Using document-level embeddings, the authors incorporate inferred skills into those extracted directly from the job ads based on their similarities.

The authors of [20] introduce Skill2vec, a technique aimed at optimizing candidate skill searches. Also inspired by Word2vec, Skill2vec maps skills into a vector space, revealing skill relationships. Training involves a neural network where skills are treated as words. This method creates a relationship graph among recruitment domain skills, and can help candidates in identifying skill gaps relative to job requirements, guiding them towards suitable training opportunities.

In [21], skill embeddings are determined using FastText [22] trained on job ads, to ensure the coherence of the extracted skills. This approach handles out-of-vocabulary instances, generating representations close to the original word, even when misspellings occur.

In [23], the author addresses the evolving Norwegian job market needs by introducing a method to identify groups of words that represent skills in the text of job listings. The use cases, requirements, data sets, implementation and design of such a skill extracting algorithm are described. The work also mentions some issues related to language ambiguity and semantic differences between datasets, which hinder precise skill extraction, and underscores the need for the computation of semantic similarities to resolve ambiguity effectively. It also suggests leveraging other NLP techniques, such as named-entity recognition (NER) and part-of-speech (POS) tagging.

A popular class of methods involve the use of supervised ML algorithms. In particular, deep neural networks have proven particularly-useful for NLP tasks. Various implementations and architectures have been developed in this direction, with promising results, often proving superior to unsupervised approaches. While methods based on deep neural networks systematically have demonstrated their reliability in capturing hidden word relationships and exhibit promising outcomes for various NLP tasks, they also have the downside of being data-intensive. Achieving favorable results in the context of skill identification demands large, labeled data sets and meticulous fine-tuning.

In [24], the authors use a LSTM architecture pre-trained to perform NER, i.e., the identification and classification of entities from unstructured text into predefined classes such as names, locations, codes, percentages, and organizations [25]. A common practice in this direction is to rely on manually labeled data sets comprising a large body of job ads, though the accuracy of manual annotation can greatly affect the reliability of the resulting models.

In [26] the authors compare models based on convolutional neural networks (CNN) and LSTM for a sentence classification problem. The CNN model incorporates word order criteria by applying a fixed-size window to the input array, consisting of words and their corresponding word embeddings [27]. The LSTM architecture takes advantage of the sequential nature of the text, addressing long-term dependencies and enabling predictions on variable-length inputs.

The authors of [28] rely on advanced language models such as Bidirectional Encoder Representations from Transformers (BERT) [29] for sentence classification in job ads. BERT is specifically designed to pre-train deep bidirectional representations from unlabeled text, considering both left and right contexts in all layers. Consequently, the pre-trained BERT model is fine-tuned with a single additional output layer, without significant modifications to the task-specific architecture. Other authors report the successful incorporation of BERT-based models into the classification pipeline. In [30], a BERT-based sentence transformer is used to perform initial feature extraction from job ad texts. Following a dimensionality reduction phase, a combination of NLP techniques and clustering methods is used to classify the job ads in the corresponding vector space. Another application of sentence transformers is by [31], who use the multilingual SBERT model [32] to determine vector representation of job ad phrases. They demonstrate that the embedding deduced by the transformer model is significantly more reliable at labeling skills in job ads than alternative unsupervised approaches, while achieving accuracies close to manual annotation.

In [33] the authors opt for a multi-label text classification approach to assign skills to each job description. Rather than classifying individual words in job descriptions, the authors treat the job descriptions as indicators for the binary relevance of multiple skills. To achieve this, they employ a BERT encoder and add an extra layer for multi-label classification. Additionally, a correlation aware bootstrapping process is introduced, encompassing structured semantic representations of skills and their co-occurrences to account for missing skills mentioned in job ads by augmenting the number of training examples.

Paper [34] addresses the scarcity of timely, comprehensive information on EU employers' skill needs by proposing a system that analyzes online job vacancies. It aims to create a pan-European platform of these vacancies for insights into skill requirements, aiming toward real-time skill demand analysis. This system employs ontologies and ML models to process multilingual job postings across various European Union languages. ML algorithms match job content to predefined terms, refining classification accuracy through expert validation and continuous ontology updates. In the proposed approach, each variable (e.g., occupation, region) and language require separate ML model training, which is currently focused on occupation classification with plans to extend to other variables.

A few survey papers provide overviews and detailed analyses of the various methods employed in the related literature for finding relevant sources of job ads in the academia [35], of knowledge extraction from job ads in the IT job market [36], or of skill identification techniques in terms of methods used, classification granularity, and existing implementations [37].

## 3. Methodology

This study aims to propose a framework for identifying meaningful relationships between employers' requirements in job ads and sets of transversal skills. To this extent, we generate an original data set consisting in job ads annotated using labels drawn from the ESCO skill base. Additionally, our goal is to create a model that can effectively predict the necessary skills for individual job descriptions. To accomplish this task, our strategy involves the implementation of a deep learning model, considering that DL neural networks, with their ability to capture complex patterns and associations within data, are well-suited for the nature of this problem. Our approach involves a comprehensive experimental study focusing on the application of neural network-based classification models for identifying transversal skills from job ads. The experimental pipeline includes several stages: initially, a preprocessing stage filters the job ads and divides them into sentences. Subsequently, a dataset generation stage creates training and test data using a sentence embedding model. Finally, the process involves generating and fine-tuning hierarchical classification models for the two ESCO skill levels.

Figure 1 summarizes the process described above.

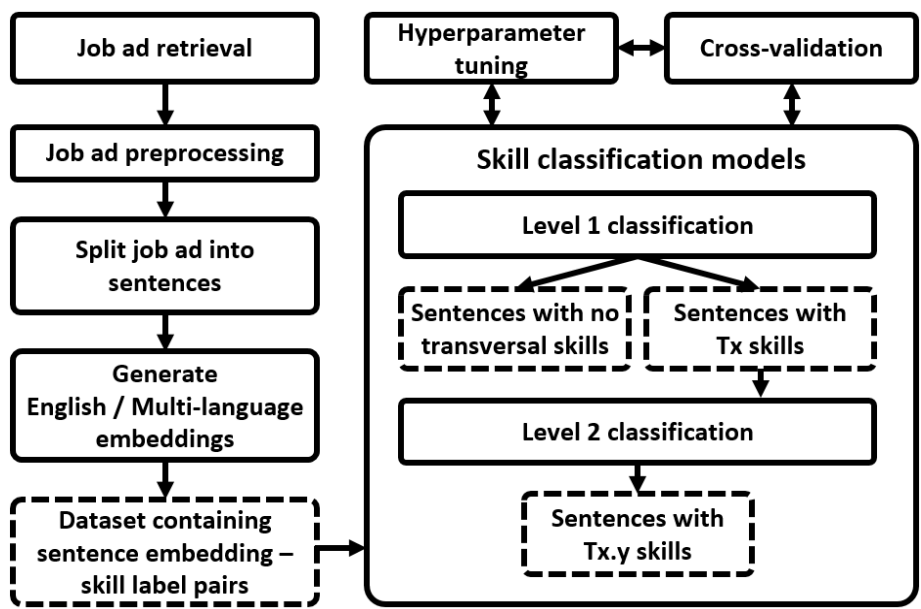

**Figure 1.** Overview of the proposed methodology

**3.1 Data Preprocessing**

The first step is the data collection and preprocessing phase, using 219 job ads downloaded from the EURES online platform [38]. We devised an automated script that parses the text, eliminates personal information, such as web addresses, and non-standard characters, and ultimately identifies the individual sentences in the ads.

Since the job ads are posted by companies from different European countries, they can be written in their respective national languages. As we will explain in the following sections, two approaches are used. The first one is to translate them into English using an automatic translation tool in order to use an English language sentence embedding model. The second one is to use a multi-language sentence embedding model applied to the original text.

These sentences then underwent manual labeling, for classifying them into skill classes and subclasses, using the skills and competences framework provided by the ESCO website [39]. ESCO, an initiative affiliated with the European Commission, serves as a classification system for the European Skills, Competences, and Occupations. The framework includes six main classes of transversal skills, each with additional subclasses:

- $T_1$ - Core skills and competences;
- $T_2$ - Thinking skills and competences;
- $T_3$ - Self-management skills and competences;
- $T_4$ - Social and communication skills and competences;
- $T_5$ - Physical and manual skills and competences;
- $T_6$ - Life skills and competences.

The utilization of ESCO's well-defined skill and competence framework ensures a structured and standardized approach to skill labeling. By classifying sentences from job ads into these skill classes, the ML models can enhance the job matching process and facilitate better alignment between job seekers and employers in the European job market.

A sample of the results of the automatic parsing and manual classification of the dataset is presented in Figure 2. Each line corresponds to a sentence. The complete set of information has several components, delimited with semicolons. The first component is an identifier of the job ad, and the second one is the identifier of the sentence in the job ad. The third component is the actual sentence in the form of a list of plain words, without punctuation marks or other identifiers. The last component includes the skills that the sentence refers to. One can notice that each sentence can belong to several classes ($T_x$) or subclasses ($T_{x,y}$). Also, there is a large number of sentences that do not represent transversal skills (marked with 0).

```
1-2; 4; Knowledge of the English language at a professional level is required; T1.1, T1.3
1-3; 1; Task description; 0
1-3; 2; Support carrying out social scientific research on social and economic aspects of environmental issues; 0
1-3; 3; Contribute to preparation of scientific outputs such as reports conference papers and journal articles; 0
1-3; 4; Manage and coordinate research work in an international context; 0
1-3; 5; Support conducting a questionnaire survey in French and in English; 0
1-3; 6; Translate scientific documents from English into French; 0
1-3; 7; Required education and experience; 0
1-3; 8; PhD degree in social sciences and humanities; 0
1-3; 9; 5 years of experience in management of scientific research preferably on an environment-related topic; 0
1-3; 10; at least ten scientific outputs such as reports conference papers journal articles books and book chapters; 0
1-3; 11; Required skills; 0
1-3; 12; Ability to manage multiple projects at the same time and to deliver them on tight Schedule; T2.3, T3.1, T3.2, T4.4
1-3; 13; Capacity to write and edit scientific reports and publications; T2.4, T3.1, T3.4, T4.5, T6.4, T6.6
1-3; 14; good communication skills suitable for teamwork in an international collaborative environment; T4.1, T4.2, T4.3, T4.4
1-3; 15; Written and oral skills as a native French speaker (preferred) and written and oral skills in English for advanced scientific communication; T1.1
1-3; 16; a very good understanding of the Canadian political and cultural context; T6.4
1-4; 1; practice in the laboratory; 0
1-4; 2; knowledge of the English language; T1.1
1-4; 3; advanced knowledge of working with a PC; T1.3
```

**Figure 2.** A sample result of data preprocessing

## 3.2 Hierarchical Skill Classification

We propose a hierarchical classification approach for the sentences extracted from the job ads. Our methodology involves several key steps aimed at enhancing the precision of skill identification and classification. In order to represent each sentence in a standardized manner, we employ a pre-trained sentence embedding model implemented in the "Sentence Transformers" library [40-41], based on the BERT architecture. This model transforms each sentence into a 768 element vector that captures its meaning. In this way, the text input is transformed into a numeric representation that can be used for further classification. The number of total sentences (i.e., instances in the data set) is 5208.

For the *first level* of classification, we separately create individual classification models using neural networks for each type of competence ($T_x$). This results in the development of six distinct models, each dedicated to determining whether or not a sentence contains a specific type of competence. This addresses the fact that a single sentence can potentially belong to multiple skill classes ($T_x$).

One of the significant challenges encountered while working with the data set was the great imbalance between sentences without skills and those containing skills. A majority of sentences in job ads typically do not contain explicit references to transversal skills. To mitigate this imbalance, we implement a data augmentation strategy; specifically, we employ sentence cloning to augment the data set by replicating the sentences that contain skills to an extent where the positive and negative samples have approximately the same number. This augmentation technique aims to improve the robustness and performance of the classification models.

An alternative to simple cloning is paraphrasing, i.e., automatically creating other sentences with different words but the same meaning for a given input sentence. This can be done, e.g., with the "Pegasus" library [42-43]. However, this procedure proved to be time consuming and the final results did not show improvements compared to the simple cloning technique. Moreover, the paraphrasing model was created only for the English language, and this posed additional limitations for working directly with multi-language models, as opposed to using English translations.

For the *second level* of the hierarchical classification, we employ a multi-label approach to further refine the classification of sentences into subclasses ($T_{x.y}$). While generating multiple single-label models is the more straightforward and easily-interpretable approach, it has the downside of not accounting for instances which belong to multiple classes, therefore oversimplifying the problem. Multi-label models have been successfully employed for text classification tasks and, while more complex and difficult to train, provide a more comprehensive representation of the relationships between classes and instances, acknowledging that instances can belong to multiple categories. This level of classification aims to provide more granular information about the specific competences referred to in the sentences. However, this stage introduces several significant challenges. The very limited number of instances available for each subclass affects both the ability to create meaningful distinct models and the generalization capabilities of the models. Also, there is a lack of clear boundaries or distinctions between some of the subclasses, i.e., the same sentence can be classified into multiple subclasses simultaneously. More specifically, we had 133 instances (i.e., sentences) in $T_1$, 238 instances in $T_2$, 378 instances in $T_3$, 395 instances in $T_4$, 20 instances in $T_5$, and 106 instances in $T_6$. Therefore, we could not train separate models for each subclass, as we did in the "level 1" classification.

In order to address these issues, we adopt a multi-label classification strategy that allows a single sentence to be assigned to multiple subclasses simultaneously. One can see some examples in Figure 3. In this case, for each sentence there are a number of $no_i$ non-mutually exclusive outputs, where $no_i$ is the number of subclasses in class $i$. For the six main classes, ($no_1$ ,…, $no_6$) = (3, 4, 4, 5, 2, 6). For example, the second line in Figure 3 contains a sentence that belongs to class $T_1$ in the first level, and to $T_{1.1}$ and $T_{1.3}$ in the second level. Since $no_1 = 3$, a binary vector of three elements defines the desired output. Since only $T_{1.1}$ and $T_{1.3}$ are relevant, the corresponding output vector is (1, 0, 1).

In the previous stages of our research, we relied on a sentence embedding model tailored for the English language to process and analyze job ads. However, the next step was to employ a *multi-language* sentence embedding model directly, to be able to effectively handle job ads written in their original languages. This broader linguistic coverage better reflects the multilingual nature of today's European job market, where job seekers and employers often interact in languages other than English.

```
Knowledge of the German language is required | ['T1.1'] | 1, 0, 0
Knowledge of the English language at a professional level is required | ['T1.1', 'T1.3'] | 1, 0, 1
Written and oral skills as a native French speaker (preferred) and written and oral skills in English for advanc
knowledge of the English language | ['T1.1'] | 1, 0, 0
advanced knowledge of working with a PC | ['T1.3'] | 0, 0, 1
Estonian and English language skills at the level necessary for work | ['T1.1'] | 1, 0, 0
Languages | ['T1.1'] | 1, 0, 0
German (A1 - Basic knowledge) | ['T1.1'] | 1, 0, 0
Language | ['T1.1'] | 1, 0, 0
English (C1 - Advanced) | ['T1.1'] | 1, 0, 0
Language English (C1 - Advanced) | ['T1.1'] | 1, 0, 0
Analyse project data with expertise and statistical modelling capability | ['T1.2', 'T1.3', 'T2.3'] | 0, 1, 1
Expertise in technology use and/or digitalization as domains and contexts of development | ['T1.3', 'T5.1', 'T6.
You have a good command in English | ['T1.1'] | 1, 0, 0
Knowledge of Luxembourgish French and/or German is an asset | ['T1.1'] | 1, 0, 0
A good command of scientific and office informatics toolsYou have a great command in English | ['T1.3', 'T2.4',
```

**Figure 3.** A sample result of data preprocessing for the "level 2" classification

## 4. Experimental Study

### 4.1 "Level 1" Classification

#### *4.1.1. The English Language Model*

We tested multiple configurations of neural networks for the classification of the vectors representing sentences. In order to assess the performance of a model, cross-validation was used. This method can assess the generalization capabilities of a ML model, ensuring that it can make accurate predictions on unseen data and avoiding overfitting, which occurs when a model is too specific to the training data and performs poorly on new, real-world examples. *k*-fold cross-validation is a widely used technique for model evaluation and selection. In this method, the dataset is divided into *k* equally sized subsets or "folds". The model is trained *k* times, each time using a different fold as the testing set and the remaining folds as the training set. This process helps to evaluate the model performance for different partitions of the data. By averaging the results from the *k* iterations, one can obtain a more reliable estimate of the model performance. *k*-fold cross-validation also provides a way to tune the hyperparameters and to assess generalization ability. In our experiments, we used *k* = 5.

    The process of training neural networks involves addressing a considerable search space encompassing network architecture, e.g., the number of hidden layers, neurons, and activation functions, as well as hyperparameters such as the choice of the optimizer (the optimization algorithm), learning rate, and regularization details. The multitude of these variables results in an overwhelming number of potential configurations, rendering exhaustive exploration unfeasible. Consequently, employing a heuristic, empirical approach becomes imperative to efficiently identify the best available options within this vast parameter space.

We explored it by iterating through various configurations, initially favoring simpler ones and progressively trying more complex architectures, and conducting multiple trials with parameter adjustments to estimate their impact.

For each configuration we repeated the training process five times, and we computed the average accuracy values for the testing sets. Although differences exist among different runs because the data are randomly selected in the cross-validation folds, they were typically less than 1% between the minimum and maximum obtained values.

In Table 1, we provide a description of the network architecture and hyperparameters that were tested, the obtained accuracy values and the primary motivation underlying the selection of each parameter combination.

In the column "Architecture and hyperparameters", the network architecture is presented on the first line in the following format:

(*number of inputs*) :
(*number of neurons in hidden layer 1*) (*activation function of the neurons in hidden layer 1*) :
… the same for the other hidden layers … :
(*number of outputs*) (*activation function of the neurons in the output layer*)
(1)

**Table 1.** Neural network configurations evaluated for the English dataset

| No. trial | Architecture and hyperparameters | Motivation |
|---|---|---|
| 1 | 768 : 128(*elu*) : 1(*sigmoid*)<br>$n = 1000$, $\eta = 0.001$, *opt* = Adam | Simple architecture (1 hidden layer) |
| 2 | 768 : 1536(*tanh*) : 512(*tanh*) : 128(*tanh*) : 32(*tanh*) : 8(*tanh*) : 1(*sigmoid*)<br>$n = 1000$, $\eta = 0.001$, *opt* = Adam | Complex architecture (5 hidden layers) |
| 3 | 768 : 20(*lrelu*) : 4(*lrelu*) : 1(*sigmoid*)<br>$n = 1000$, $\eta = 0.001$, *opt* = Adam | Medium size architecture (2 hidden layers), strong information compression in the first hidden layer |
| 4 | 768 : 81(*lrelu*) : 9(*lrelu*) : 1(*sigmoid*)<br>$n = 1000$, $\eta = 0.001$, *opt* = Adam | Medium size architecture (2 hidden layers), balanced information compression |
| 5 | 768 : 81(*sigmoid*) : 9(*sigmoid*) : 1(*sigmoid*)<br>$n = 1000$, $\eta = 0.001$, *opt* = Adam | Sigmoid activation function in the hidden layers |
| 6 | 768 : 81(*tanh*) : 9(*tanh*) : 1(*sigmoid*)<br>$n = 1000$, $\eta = 0.001$, *opt* = Adam | Hyperbolic tangent activation function in the hidden layers |
| 7 | 768 : 81(*lrelu*) : 9(*lrelu*) : 1(*sigmoid*)<br>$n = 1000$, $\eta = 0.001$, *opt* = RMSprop | Leaky ReLU activation function in the hidden layers and the RMSprop optimizer |
| 8 | 768 : 81(*lrelu*) : 9(*lrelu*) : 1(*sigmoid*)<br>$n = 1000$, $\eta = 0.01$, *opt* = Adam | Higher learning rate (0.01) |
| 9 | 768 : 81(*lrelu*) : 9(*lrelu*) : 1(*sigmoid*)<br>$n = 10000$, $\eta = 0.01$, *opt* = Adam | Large number of training epochs (10000) |
| 10 | 768 : 81(*lrelu*) : 9(*lrelu*) : 1(*sigmoid*)<br>$n = 100$, $\eta = 0.001$, *opt* = Adam | Small number of training epochs (100) |

The activation functions that we used are: *sigmoid* (the unipolar sigmoid), *tanh* (the hyperbolic tangent), *elu* (exponential linear unit), and *lrelu* (leaky rectified linear unit).

Then, the main hyperparameters are mentioned on the second line: $n$ is the number of training epochs, $\eta$ is the learning rate, and *opt* represents the optimization algorithm.

For all configurations, the binary cross-entropy loss function was used, as it is better-suited to classification problems.

The best results were obtained with balanced configurations, i.e., which are sufficiently complex to capture the underlying patterns of the data but without excessive complexity that may lead to overfitting. This equilibrium extends to both the architectural complexity of the network and the number of training epochs.

Furthermore, it is important to note that the best model varies between the different classes of skills ($T_x$), and therefore different neural models are used in each specific case.

Figure 4 shows the best cross-validation results for the six "level 1" models corresponding to the six main skill classes ($T_x$). One can see that the accuracy values for all models are quite high, above 94%, with the best one reaching 99%. We should emphasize that these values represent the averages obtained for the testing sets, not for the training sets. Therefore, we can conclude that the "level 1" classification models are well suited for the given task.

*4.1.2. The Multi-language Model*

Our evaluation revealed that the performance results achieved using the multi-language model were remarkably close to those obtained when using the English-specific model. In this subsection we show the results for the "level 1" problem, but the hierarchical classification methodology designed here is independent of the specific sentence embedding model. The network architectures and hyperparameters that were tested for the multi-language sentence embeddings are presented in Table 2.

**Table 2.** Neural network configurations evaluated for the multi-language dataset

| No. trial | Architecture and hyperparameters | Motivation |
|---|---|---|
| 1 | 768 : 128(*elu*) : 1(*sigmoid*) <br> $n = 1000$, $\eta = 0.001$, *opt* = Adam | Simple architecture (1 hidden layer) |
| 2 | 768 : 128(*elu*) : 1(*sigmoid*) <br> $n = 1000$, $\eta = 0.01$, *opt* = Adam | Higher learning rate (0.01) |
| 3 | 768 : 128(*elu*) : 1(*sigmoid*) <br> $n = 1000$, $\eta = 0.01$, *opt* = RMSprop | RMSprop optimizer |
| 4 | 768 : 128(*elu*) : 1(*sigmoid*) <br> $n = 10000$, $\eta = 0.01$, *opt* = Adam | Large number of training epochs (10000) |
| 5 | 768 : 128(*lrelu*) : 1(*sigmoid*) <br> $n = 1000$, $\eta = 0.01$, *opt* = Adam | Leaky ReLU activation function in the hidden layer |
| 6 | 768 : 81(*lrelu*) : 9(*lrelu*) : 1(*sigmoid*) <br> $n = 1000$, $\eta = 0.01$, *opt* = Adam | Medium size architecture (2 hidden layers), balanced information compression |
| 7 | 768 : 81(*lrelu*) : 9(*lrelu*) : 1(*sigmoid*) <br> $n = 1000$, $\eta = 0.01$, *opt* = RMSprop | RMSprop optimizer |

The results of the experiments, in terms of accuracy, are presented in Table 3. The "Trial" column corresponds to the configurations in Tables 1 and 2. The rest of the columns show the accuracy values obtained for the testing sets in the cross-validation procedure. The best results for each skill class are highlighted in bold red. Still, one can see that several models can give comparably good results. Therefore, the results that are within a 0.2% range from the best one are also marked in bold italics.

**Table 3.** Accuracy values obtained for the evaluated neural network configurations for the first level classification

| *English language model* | | | | | | |
|---|---|---|---|---|---|---|
| **Trial** | **T1** | **T2** | **T3** | **T4** | **T5** | **T6** |
| 1 | 96.91 | 95.16 | 92.70 | 93.68 | 99.33 | 96.14 |
| 2 | 95.93 | 93.55 | 91.74 | 93.26 | 98.52 | 95.14 |
| 3 | 97.20 | 94.87 | 93.03 | 94.18 | 99.37 | 96.76 |
| 4 | **97.50** | 95.85 | 93.59 | 94.60 | *99.52* | *97.25* |
| 5 | 97.10 | 94.68 | 91.99 | 93.76 | 99.17 | 95.62 |
| 6 | 96.91 | 94.68 | 92.34 | 93.28 | 99.29 | 95.80 |
| 7 | *97.45* | 95.91 | 93.59 | *94.78* | *99.54* | *97.33* |
| 8 | *97.48* | *96.04* | 93.74 | **94.84** | *99.50* | 97.14 |
| 9 | *97.48* | **96.24** | **94.07** | 94.43 | **99.58** | **97.43** |
| 10 | 95.35 | 90.00 | 89.55 | 89.19 | 97.81 | 91.03 |

| *Multi-language model* | | | | | | |
|---|---|---|---|---|---|---|
| **Trial** | **T1** | **T2** | **T3** | **T4** | **T5** | **T6** |
| 1 | 97.14 | 94.71 | 92.92 | 93.26 | 98.93 | 96.01 |
| 2 | 97.54 | 89.11 | 93.66 | 94.08 | 99.03 | 96.95 |
| 3 | 97.79 | 95.61 | 93.80 | 94.69 | *99.20* | 96.70 |
| 4 | 97.56 | *95.86* | *93.97* | 94.67 | 99.04 | 96.97 |
| 5 | *97.82* | **96.01** | 93.89 | *94.83* | 99.27 | *97.23* |
| 6 | **98.02** | 95.82 | *94.12* | **94.90** | 99.24 | 97.08 |
| 7 | *97.98* | *95.97* | *94.04* | 94.56 | **99.39** | 97.02 |

As shown in Figures 4 and 5, the multi-language model has very close accuracy values compared to the English language model. In Figure 5, we also include the relative difference between the two models computed with the following equation:

$$r_d = \frac{a_m}{a_e} - 1 \qquad (2)$$

where $a_m$ is the best accuracy obtained with the multi-language embeddings and $a_e$ is the best accuracy obtained with the English language embeddings.

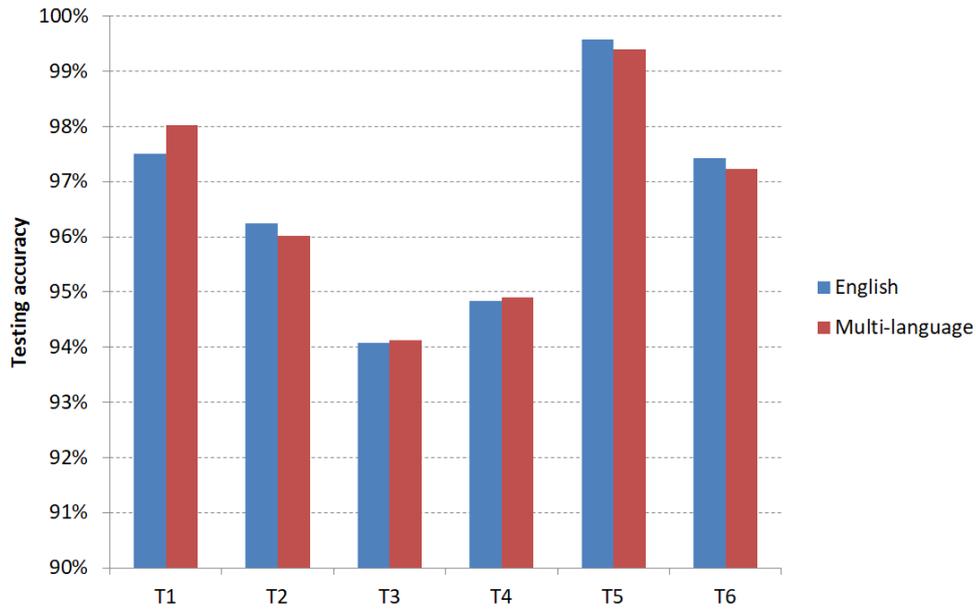

**Figure 4.** Comparison between the performance of classification using English language and multi-language sentence embeddings

One can see that the multi-language model is slightly better for the classes $T_1$, $T_3$ and $T_4$, and slightly worse for the rest, but the accuracy values are all within a 0.55% range. Given the approximate nature of neural network-based classification, we can conclude that the quality of the results obtained in the two situations is basically the same. However, the multi-language model has the advantage of flexibility and using it avoids the additional step of detecting the original language and performing an automatic translation that may even distort the original message to some extent. Therefore, it was selected as the default model to be used for the classification of skills, and also for the results of the second level classification presented in the next subsection.

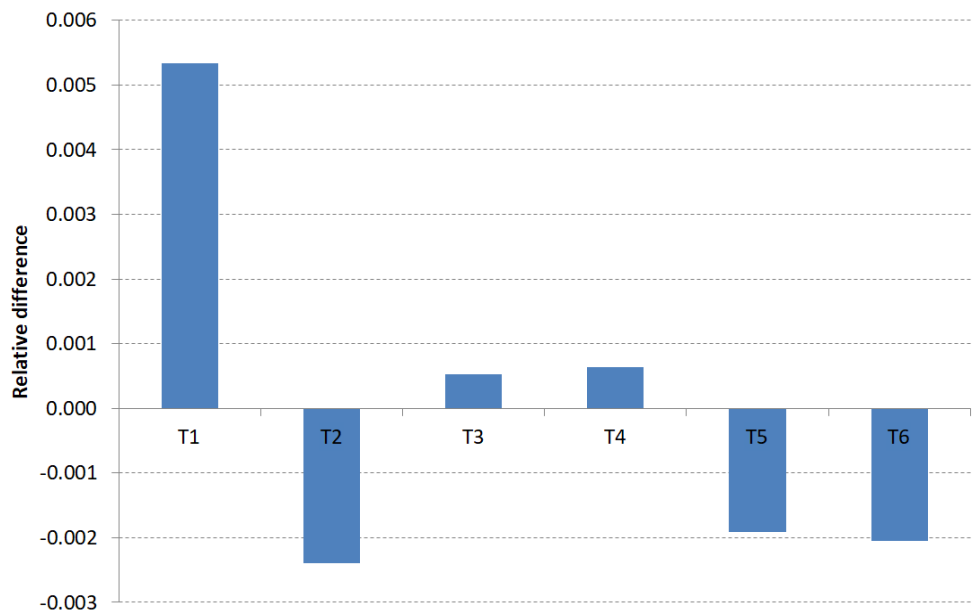

**Figure 5.** Relative difference obtained by the multi-language model compared to the English language model

## 4.2. "Level 2" Classification

A similar methodology was adopted for the "level 2" classification. In all cases, a variable, class-specific number of neurons were used in the output layer, denoted as *no*. Because the small number of training instances may lead to overfitting, we also tested the effect of regularization. The corresponding factor, denoted as *λ*, represents the regularization rate.

The evaluated neural network configurations are presented in Table 4.

**Table 4.** Neural network configurations evaluated for the second level classification

| No. trial | Architecture and hyperparameters | Motivation |
|---|---|---|
| 1 | 768 : 128(*lrelu*) : *no*(*sigmoid*) <br> $n = 100$, $\eta = 0.001$, *opt* = Adam | Simple architecture (1 hidden layer) |
| 2 | 768 : 128(*lrelu*) : *no*(*sigmoid*) <br> $n = 100$, $\eta = 0.01$, *opt* = Adam | Higher learning rate (0.01) |
| 3 | 768 : 150(*lrelu*) : 30(*lrelu*) : *no*(*sigmoid*) <br> $n = 100$, $\eta = 0.01$, *opt* = Adam | Larger architecture (2 hidden layers) |
| 4 | 768 : 128(*lrelu*) : *no*(*sigmoid*) <br> $n = 100$, $\eta = 0.01$, $\lambda = 10^{-5}$, *opt* = Adam | Regularization |
| 5 | 768 : 128(*lrelu*) : *no*(*sigmoid*) <br> $n = 1000$, $\eta = 0.01$, $\lambda = 10^{-5}$, *opt* = Adam | Larger number of training epochs (1000) |
| 6 | 768 : 128(*tanh*) : *no*(*sigmoid*) <br> $n = 100$, $\eta = 0.01$, $\lambda = 10^{-5}$, *opt* = Adam | Hyperbolic tangent activation function in the hidden layer |
| 7 | 768 : 128(*lrelu*) : *no*(*sigmoid*) <br> $n = 50$, $\eta = 0.01$, $\lambda = 10^{-5}$, *opt* = Adam | Smaller number of training epochs (50) |
| 8 | 768 : 128(*elu*) : *no*(*sigmoid*) <br> $n = 100$, $\eta = 0.01$, $\lambda = 10^{-5}$, *opt* = Adam | Exponential Linear Unit (ELU) activation function in the hidden layer |
| 9 | 768 : 128(*elu*) : *no*(*sigmoid*) <br> $n = 100$, $\eta = 0.01$, $\lambda = 10^{-5}$, *opt* = RMSprop | RMSprop optimizer |
| 10 | 768 : 150(*sigmoid*) : 30(*sigmoid*) : *no*(*sigmoid*) <br> $n = 100$, $\eta = 0.01$, $\lambda = 10^{-5}$, *opt* = Adam | Larger architecture (2 hidden layers) with sigmoid activation function in the hidden layers |

Table 5 presents the results of the experiments for the second level of classification. The same conventions regarding colors and font styles as in Table 4 are used: bold red denotes the best results, while bold italics represent results within 0.2% proximity to the best. Unlike the results for the first level, the accuracy values for cross-validation are lower here. The results for the training sets are not included, but they were all above 90%, and the general average across all trials and all skill subclasses was 95.5%.

**Table 5.** Accuracy values obtained for the evaluated neural network configurations for the second level classification

| Trial | T1 | T2 | T3 | T4 | T5 | T6 |
|---|---|---|---|---|---|---|
| 1 | 91.46 | 66.93 | 72.23 | 74.49 | 76.20 | 81.55 |
| 2 | 92.95 | 67.69 | **73.09** | *76.05* | 77.47 | 82.51 |
| 3 | 88.81 | 65.41 | 69.83 | 72.91 | 73.91 | 78.61 |
| 4 | 95.89 | **68.17** | *73.05* | 75.23 | 75.67 | **84.82** |
| 5 | 92.82 | 65.47 | 69.61 | 73.34 | 77.13 | 81.07 |
| 6 | *96.22* | 67.53 | 72.44 | *76.05* | **80.05** | 84.32 |
| 7 | 93.19 | 65.63 | 70.40 | 73.34 | 77.68 | 81.92 |
| 8 | 91.71 | 65.09 | 69.34 | 72.67 | 76.87 | 81.03 |
| 9 | 95.02 | 66.98 | 71.46 | 74.72 | 78.79 | 83.02 |
| 10 | *96.35* | 67.85 | 72.16 | **76.21** | 76.75 | 83.10 |

The average training and testing results in terms of accuracy, for the $k = 5$ cross-validation folds are presented in Figure 6. Since this is a multi-label classification, the accuracy value for an instance represents the average of the accuracy values obtained for all outputs.

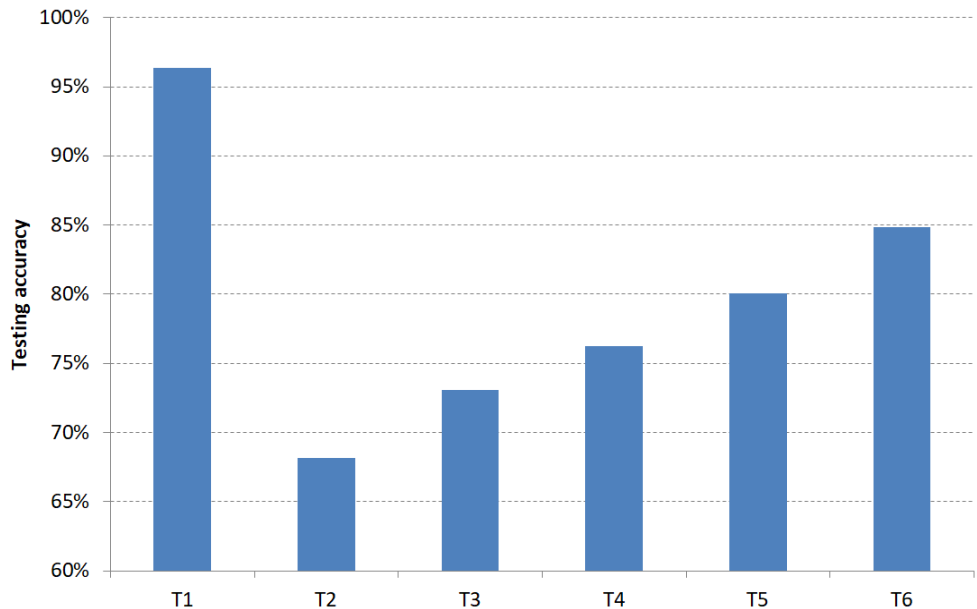

**Figure 6.** The results obtained for the "level 2" classification

In this case, one can notice that the testing performance is not as good as for the "level 1" task. It is likely that the root causes are the low number of instances, presented in Table 6, and the ambiguity of the subclass membership.

Table 6. The total number of instances in each subclass

| T1 | T2 | T3 | T4 | T5 | T6 |
|---|---|---|---|---|---|
| T1.1: 106 | T2.1: 99 | T3.1: 186 | T4.1: 169 | T5.1: 8 | T6.1: 8 |
| T1.2: 7 | T2.2: 101 | T3.2: 176 | T4.2: 181 | T5.2: 13 | T6.2: 10 |
| T1.3: 29 | T2.3: 137 | T3.3: 94 | T4.3: 209 | | T6.3: 15 |
| | T2.4: 107 | T3.4: 122 | T4.4: 142 | | T6.4: 31 |
| | | | T4.5: 54 | | T6.5: 32 |
| | | | | | T6.6: 33 |

The experimental results show that our models generally perform well in identifying transversal skills within job ads, considering the limited available data. In particular, the high accuracies of the "level 1" models allow the reliable screening of job ads that are lacking in phrases containing transversal skills, which we found to be in the vast majority. Identifying individual transversal skills requires data in an amount that can sufficiently cover the space of possible skill-related phrasings. Currently, our "level 2" models achieve high accuracies for the T1 category, which focuses on language skills. In this case, the phrasings are generally consistent (eg. "Has good knowledge of [language]" or "A good grasp of [language] is beneficial"). High accuracy is also achieved for the T6 category, where, similarly, the phrasings found in job ads are more consistent than for other categories. In most cases, some of the best results were obtained using neural networks consisting in two hidden layers with descending sizes. We found that extending the architecture with additional layers of greater sizes did not improve the accuracy, instead resulting in overcomplicated models. However, especially in the case of "level 2" models, we surmise that a far greater improvement would be achieved with a broader data set with better coverage of possible phrasings, than by tweaking our current models.

## 5. Conclusions

The present study attempts to create a robust framework for defining the correlations between the requirements of job advertisements and transversal skills, in order to predict the required skills for individual job descriptions. This is achieved by means of a hierarchical classification methodology, utilizing ESCO taxonomy. The comparison between results obtained with English-specific and multi-language sentence embeddings reveals comparable performance, validating the adaptability and efficiency of multi-language embeddings, subsequently adopted as the default model due to its flexibility. At the forefront of our methodology lie the "level 1" and "level 2" hierarchical classifications, each showcasing high accuracy, yet "level 2" exhibits lower cross-validation accuracy attributed to subclass ambiguity and limited number of instances.

The main contributions of this approach consist in developing neural-network models to classify job ads within the hierarchical skill framework of the European ESCO taxonomy. The key innovations include a self-generated dataset comprising manually labeled job ads referencing specific skills. Their classification aligns with ESCO hierarchical structure, distinguishing between broader top-level skills and more specialized subskills.

Future research directions may explore refining subclass identification methods to tackle ambiguity issues and improve classification accuracy. Augmenting data instances within subclasses and employing advanced data augmentation techniques could enhance model generalization. Investigating multi-task learning approaches, enabling simultaneous classification of multiple subclasses, could deepen understanding and granularity in skill identification. Exploring domain adaptation techniques to address linguistic variations across job descriptions in different languages remains very important. Additionally, investigating interpretable models or explainable AI methodologies can aid in understanding model decisions, fostering trust and applicability in real-world scenarios. Collaborations with industry partners for larger datasets and validation in diverse job markets would further validate the efficacy of the proposed framework.

## Acknowledgement

This work was supported by the DocTalent4EU project – Transforming Europe through Doctoral Talent and Skills Recognition - Horizon-Widera, GA no. 101095292, 2022-2024.

## References

1. W. Zhou, Y. Zhu, F. Javed, M. Rahman, J. Balaji, and M. McNair, "Quantifying skill relevance to job titles," in 2016 IEEE International Conference on Big Data (Big Data), pp. 1532–1541, 2016

2. M. Papoutsoglou, N. Mittas, and L. Angelis, "Mining People Analytics from StackOverflow Job Advertisements," in 2017 43rd Euromicro Conference on Software Engineering and Advanced Applications (SEAA), pp. 108–115, 2017.

3. E. Malherbe and M. A. Aufaure, "Bridge the terminology gap between recruiters and candidates: A multilingual skills base built from social media and linked data," in 2016 IEEE/ACM International Conference on Advances in Social Networks Analysis and Mining (ASONAM), pp. 583–590, 2016.

4. E. M. Sibarani, S. Scerri, C. Morales, S. Auer, and D. Collarana, "Ontology-guided Job Market Demand Analysis: A Cross-Sectional Study for the Data Science field," in Proceedings of the 13th International Conference on Semantic Systems, ser. Semantics2017. New York, NY, USA: Association for Computing Machinery, pp. 25–32, 2017.

5. A. Gardiner, C. Aasheim, P. Rutner, and S.Williams, "Skill Requirements in Big Data: A Content Analysis of Job Advertisements," Journal of Computer Information Systems, vol. 58, no. 4, pp. 374–384, 2018.

6. H. Chaibate, A. Hadek, S. Ajana, S. Bakkali, and K. Faraj, "Analyzing the engineering soft skills required by Moroccan job market," in 2019 5th International Conference on Optimization and Applications (ICOA), pp. 1–6, 2019.

7. F. Niederman and M. Sumner, "Resolving the Skills Paradox: A Content Analysis of a Jobs Database," in Proceedings of the 2019 on Computers and People Research Conference, pp. 164–167, 2019.

8. J. A. Rios, G. Ling, R. Pugh, D. Becker, and A. Bacall, "Identifying Critical 21st-Century Skills for Workplace Success: A Content Analysis of Job Advertisements," Educational Researcher, vol. 49, no. 2, pp. 80–89, 2020.

9. S. Debortoli, O. Maller, and J. v. Brocke, "Comparing Business Intelligence and Big Data Skills," Business & Information Systems Engineering, vol. 6, no. 5, pp. 289–300, 2014.

10. A. De Mauro, M. Greco, M. Grimaldi, and P. Ritala, "Human resources for Big Data professions: A systematic classification of job roles and required skill sets," Information Processing & Management, vol. 54, no. 5, pp. 807–817, 2018.

11. F. Gurcan and S. Sevik, "Expertise Roles and Skills Required by the Software Development Industry," in 2019 1st International Informatics and Software Engineering Conference (UBMYK), pp. 1–4, 2019.

12. F. Gurcan and N. E. Cagiltay, "Big Data Software Engineering: Analysis of Knowledge Domains and Skill Sets Using LDA-Based Topic Modeling," IEEE Access, vol. 7, pp. 82 541–82 552, 2019.

13. H. Jelodar, Y. Wang, C. Yuan, X. Feng, X. Jiang, Y. Li, and L. Zhao, "Latent Dirichlet allocation (LDA) and topic modeling: models, applications, a survey," Multimedia Tools and Applications, vol. 78, no. 11, pp. 15 169–15 211, 2019.

14. F. Javed, P. Hoang, T. Mahoney, and M. McNair, "Large-scale occupational skills normalization for online recruitment," in Twenty-Ninth IAAI Conference, 2017.

15. M. Zhao, F. Javed, F. Jacob, and M. McNair, "SKILL: A System for Skill Identification and Normalization," in Proceedings of the Twenty-Ninth AAAI Conference on Artificial Intelligence, pp. 4012–4017, 2015.

16. T. Mikolov, I. Sutskever, K. Chen, G. S. Corrado, and J. Dean, "Distributed representations of words and phrases and their compositionality," in Advances in neural information processing systems, pp. 3111–3119, 2013.

17. D. Botov, J. Klenin, A. Melnikov, Y. Dmitrin, I. Nikolaev, and M. Vinel, "Mining Labor Market Requirements Using Distributional Semantic Models and Deep Learning," in Business Information Systems, ser. Lecture Notes in Business Information Processing, W. Abramowicz and R. Corchuelo, Eds. Cham: Springer International Publishing, 2019, pp. 177–190.

18. A. Gugnani and H. Misra, "Implicit Skills Extraction Using Document Embedding and Its Use in Job Recommendation." in AAAI, pp. 13 286–13 293, 2020.

19. Q. Le and T. Mikolov, "Distributed representations of sentences and documents," in International conference on machine learning. PMLR, pp. 1188–1196, 2015.

20. Le Van-Duyet and Vo Minh Quan and Dang Quang An, Skill2vec: Machine Learning Approach for Determining the Relevant Skills from Job Description, 2019, https://arxiv.org/abs/1707.09751v3.

21. S. Li, B. Shi, J. Yang, J. Yan, S. Wang, F. Chen, and Q. He, "Deep Job Understanding at LinkedIn," in Proceedings of the 43rd International ACM SIGIR Conference on Research and Development in Information Retrieval, ser. SIGIR '20. New York, NY, USA: Association for Computing Machinery, 2020, pp. 2145–2148.

22. P. Bojanowski, E. Grave, A. Joulin, and T. Mikolov, "Enriching Word Vectors with Subword Information," Transactions of the Association for Computational Linguistics, vol. 5, pp. 135–146, 2017.

23. A. G. Fagerbakk, "Keeping Up with the Market: Extracting competencies from Norwegian job listings", Master's thesis, The Arctic University of Norway, 2021.

24. S. Jia, X. Liu, P. Zhao, C. Liu, L. Sun, and T. Peng, "Representation of Job-Skill in Artificial Intelligence with Knowledge Graph Analysis," in 2018 IEEE Symposium on Product Compliance Engineering - Asia (ISPCE-CN), pp. 1–6, 2018.

25. G. Lample, M. Ballesteros, S. Subramanian, K. Kawakami, and C. Dyer, "Neural Architectures for Named Entity Recognition," in Proceedings of the 2016 Conference of the North American Chapter of the Association for Computational Linguistics: Human Language Technologies, pp. 260–270, 2016.

26. L. Sayfullina, E. Malmi, and J. Kannala, "Learning Representations for Soft Skill Matching," in Analysis of Images, Social Networks and Texts, Lecture Notes in Computer Science, Springer International Publishing, pp. 141–152, 2018.

27. Y. Kim, "Convolutional Neural Networks for Sentence Classification," in Proceedings of the 2014 Conference on Empirical Methods in Natural Language Processing (EMNLP), pp. 1746–1751, 2014.

28. D. A. Tamburri, W.-J. V. D. Heuvel, and M. Garriga, "DataOps for Societal Intelligence: a Data Pipeline for Labor Market Skills Extraction and Matching," in 2020 IEEE 21st International Conference on Information Reuse and Integration for Data Science (IRI), pp. 391–394, 2020.

29. J. Devlin, M.-W. Chang, K. Lee, and K. Toutanova, "BERT: Pre-training of Deep Bidirectional Transformers for Language Understanding," arXiv:1810.04805 [cs], May 2019.

30. M. Lukauskas, V. Šarkauskaitė, V. Pilinkienė, A. Stundžienė, A. Grybauskas, and J. Bruneckienė, "Enhancing Skills Demand Understanding through Job Ad Segmentation Using NLP and Clustering Techniques," Applied Sciences, vol. 13, no. 10, p. 6119, 2023.

31. M. Mathiasen, J. Nielsen, and S. Laub, "A Transformer Based Semantic Analysis of (non-English) Danish Job ads," Proceedings of the 15th International Conference on Computer Supported Education, 2023.

32. N. Reimers and I. Gurevych, "Sentence-BERT: Sentence Embeddings using Siamese BERT-Networks," Proceedings of the 2019 Conference on Empirical Methods in Natural Language Processing and the 9th International Joint Conference on Natural Language Processing (EMNLP-IJCNLP), 2019.

33. A. Bhola, K. Halder, A. Prasad, and M.-Y. Kan, "Retrieving Skills from Job Descriptions: A Language Model Based Extreme Multi-label Classification Framework," in Proceedings of the 28th International Conference on Computational Linguistics, pp. 5832–5842, 2020.

34. Cedefop (2019). "Online job vacancies and skills analysis: a Cedefop pan-European approach". Luxembourg: Publications Office. http://data.europa.eu/doi/10.2801/097022.

35. R. Applegate, "Job ads, jobs, and researchers: Searching for valid sources," Library & Information Science Research, vol. 32, no. 2, pp. 163–170, 2010.

36. M. Papoutsoglou, A. Ampatzoglou, N. Mittas, and L. Angelis, "Extracting Knowledge From On-Line Sources for Software Engineering Labor Market: A Mapping Study," IEEE Access, vol. 7, pp. 157 595–157 613, 2019.

37. I. Khaouja, I. Kassou, and M. Ghogho, "A Survey on Skill Identification From Online Job Ads," IEEE Access, vol. 9, pp. 118134–118153, 2021.

38. European Commission, Directorate-General for Employment, Social Affairs and Inclusion, EURES, 2023, online https://eures.ec.europa.eu/index_en.

39. European Commission, Directorate-General for Employment, Social Affairs and Inclusion, ESCO Publications: Skills & competences, 2022, available online: https://esco.ec.europa.eu/en/classification/skill_main.

40. N. Reimers and I. Gurevych, "Sentence-BERT software library", 2022, online: https://www.sbert.net.

41. N. Reimers and I. Gurevych, "Making Monolingual Sentence Embeddings Multilingual using Knowledge Distillation", Proceedings of the 2020 Conference on Empirical Methods in Natural Language Processing, 2020, Association for Computational Linguistics.

42. J. Zhang, Y. Zhao, M. Saleh and P. J. Liu, "PEGASUS: Pre-training with Extracted Gap-sentences for Abstractive Summarization", 2020, paper: https://arxiv.org/abs/1912.08777v3; software library available online: https://github.com/google-research/pegasus.

43. A. Adarsh, "Pegasus-paraphrase library", 2022, online: https://github.com/adarshgowdaa/pegasus-paraphrase.